\documentclass[conference]{IEEEtran}
\IEEEoverridecommandlockouts
\usepackage{cite}
\usepackage{amsmath,amssymb,amsfonts}
\usepackage{algorithmic}
\usepackage{graphicx}
\usepackage{textcomp}
\usepackage{xcolor}
\usepackage{orcidlink}
\usepackage{hyperref}
\usepackage{balance}
\usepackage{romanbar} 
\usepackage[dvipsnames]{xcolor} 
\def\BibTeX{{\rm B\kern-.05em{\sc i\kern-.025em b}\kern-.08em
    T\kern-.1667em\lower.7ex\hbox{E}\kern-.125emX}}
\begin{document}

\title{Towards Airborne Object Detection: A Deep Learning Analysis
	\\
{
}
}

\author{\IEEEauthorblockN{1\textsuperscript{st} Prosenjit Chatterjee$^\textsuperscript{\orcidlink{0000-0003-1169-4717}}$}
\IEEEauthorblockA{\textit{Member, IEEE} \\
\href{https://orcid.org/0000-0003-1169-4717}{0000-0003-1169-4717} \\
}

\and
\IEEEauthorblockN{2\textsuperscript{nd} ANK Zaman$^\textsuperscript{\orcidlink{0000-0001-7831-0955}}$}
\IEEEauthorblockA{\textit{Member, IEEE} \\
\href{https://orcid.org/0000-0001-7831-0955}{0000-0001-7831-0955} \\
}
    }

\maketitle
\begin{abstract}
The rapid proliferation of airborne platforms—including commercial aircraft, drones, and UAVs—has intensified the need for real‑time, automated threat assessment systems. Current approaches depend heavily on manual monitoring, resulting in limited scalability and operational inefficiencies. This work introduces a dual‑task model based on EfficientNetB4 capable of performing airborne object classification and threat‑level prediction simultaneously. To address the scarcity of clean, balanced training data, we constructed the AODTA Dataset by aggregating and refining multiple public sources. We benchmarked our approach on both the AVD Dataset and the newly developed AODTA Dataset, and further compared performance against a ResNet‑50 baseline, which consistently underperformed relative to EfficientNetB4. Our EfficientNetB4 model achieved 96\% accuracy in object classification and 90\% accuracy in threat‑level prediction, underscoring its promise for applications in surveillance, defense, and airspace management. Although the title references “detection,” this study focuses specifically on classification and threat‑level inference using pre‑localized airborne object images provided by existing datasets.
\end{abstract}


\section{Introduction}\label{sec:s1}
Airborne object detection and classification are essential in defense, surveillance, and aviation management, where accurate object recognition and real-time threat assessment are critical for effective response. This research focuses on a dual-task neural network model based on EfficientNetB4~\cite{TensorFlow_EfficientNetB4}, designed to perform both airborne object classification and threat-level prediction. A major contribution of this work is the introduction of a new AODTA Dataset~\cite{chatterjee2025airborne}—the first curated dataset combining multiple public sources to provide clean, diverse, and well-balanced airborne imagery. Our results show that EfficientNetB4 performs significantly better than ResNet-50, particularly on the AODTA Dataset~\cite{chatterjee2025airborne}, achieving high accuracy and reliable performance. Together, the proposed dataset and model offer a practical and effective approach to improving airborne object detection and threat analysis.

It is essential to clarify that our study focuses on the \emph{classification and threat-level prediction} of airborne objects rather than End-to-end detection. The datasets used (AVD and AODTA) already provide cropped airborne object images. Therefore, our model does not perform object localization, but instead classifies objects into four categories (airplane, drone, helicopter, UAV) and simultaneously predicts their threat level (low, medium, high). This scope complements existing object detection pipelines, and our framework can be integrated with detectors such as YOLO or Faster R-CNN in future work.

The remainder of the paper is organized as follows: Section~\ref{sec:s2} discusses literature review, Section~\ref{sec:s3} data collection and preprocessing, Section~\ref{sec:s4} describes the model architecture, Section~\ref{sec:s5} presents the obtained results, discussions, and analysis, and finally Section~\ref{sec:s6} provides the concluding remarks.

\section{Literature Review}\label{sec:s2}
Automated image recognition has advanced substantially within security and surveillance applications, with many studies highlighting the utility of machine learning (ML) in object classification. Existing studies have explored the architectures of YOLO~\cite{klingler2022yolov3}, ResNet~\cite{torchvision-resnet50-docs}, and VGGNet~\cite{boesch2022vggnet} for aerial image classification. But despite these developments, a few studies specifically address multi-tiered threat-level assessment for airborne objects. 

This research builds on established methods while focusing on comprehensive threat classification using custom-curated datasets of airborne images. Existing research has explored techniques such as YOLO~\cite{klingler2022yolov3}, ResNet~\cite{torchvision-resnet50-docs}, and VGGNet~\cite{boesch2022vggnet} for image classification. Some of the recently published works are discussed below.

Ensuring drone safety requires early detection of fast-moving aggressive birds, a task deep learning has recently enabled using RGB imagery~\cite{9511386}. However, the lack of drone-captured bird detection datasets limits progress in this domain. Fujii et al.~\cite{9511386} introduce a novel dataset comprising 34,467 bird instances across 21,837 images collected under diverse conditions. Experimental evaluations reveal that even state-of-the-art (SOTA) models struggle with this challenging dataset. Additionally, the study shows that while common data augmentation techniques may be ineffective, selectively chosen methods can enhance detection performance.

Airborne object detection and classification are critical for real-time applications in surveillance and public safety. Puduru et al.~\cite{10968603} utilizes the AOD 4 dataset comprising 22,516 images under varied environmental conditions, categorized into four classes: airplanes, helicopters, drones, and birds. A custom lightweight CNN is developed, achieving a high classification accuracy of 97.12\%, outperforming several state-of-the-art pre-trained models. The model demonstrates real-time performance with an inference time of 2.9 ms on an Intel Core i7, marking a 4× speedup over MobileNetv3Small. These results highlight the model’s suitability for efficient airborne threat detection in practical scenarios.

Choi and Jo~\cite{9589099} focuses on ground object detection and classification in drone imagery, where objects appear small and vary significantly due to perspective and altitude changes. An attention-based CNN architecture is proposed to address these challenges, achieving a precision of 87.12\% on a curated drone image dataset. The model outperforms baseline networks such as MobileNet, VGG16, SqueezeNet, and ResNet in both accuracy and speed. It operates nearly three times faster than VGG16 and twice as fast as MobileNet, with a more compact parameter footprint. These findings demonstrate the effectiveness of attention mechanisms for real-time drone-based surveillance and object tracking.

With the rising accessibility of drones, there is a growing need for systems that prevent unauthorized aerial intrusions. Gracia et al.~\cite{9289397} presents a visual sensing-based anti-drone system utilizing Faster R-CNN with a ResNet-101 backbone. The model is trained on the SafeShore project dataset, targeting drone detection. Achieving an accuracy of 93.40\%, the system effectively identifies drones in simulation environments. The results highlight the potential of deep learning-based object detection for reliable drone surveillance and airspace security.

However, most methods are limited to single-task classification, neglecting multi-tiered threat-level assessment. Key references include Koutsoubis et al.~\cite{koutsoubis2023drone}, who introduced a machine-learning-based drone detection system focusing on explainable AI and dataset development, emphasizing transparency and trust in predictions~\cite{Arun_Dua_2024}, extended classification methods to aerial objects, highlighting risk-based classification but lacked multi-output capabilities. Mrabet et al.~\cite{Mrabet_Sliti_Ammar_2024} presented drone detection algorithms, revealing a gap in integrating threat-level assessments with real-world applications. Du et al. (2021) displayed the vulnerability of deep neural networks in potential threats via satellite imagery processing ~\cite{du2021physicaladversarialattacksaerial}. Showcasing the need for robust models in aerial threat analysis. Qu et al.~\cite{qu2024intention} introduced three deep learning recognition models to detect intentions of air target operations after model training. While these contributions laid the groundwork for robust detection systems, the novelty of this work lies in simultaneously addressing class and threat-level prediction.

Xu et al.~\cite{XU2024103732} experimented with deep learning techniques use for detecting animals using aerial and satellite imagery. By using models such as YOLO~\cite{bochkovskiy2020yolov4}, Faster R-CNN, and U-Net their research was shown to show proficiency in object detection and segmentation~\cite{XU2024103732}. Lin et al. integrated EfficientNet-B4~\cite{TensorFlow_EfficientNetB4} for precise feature extraction, combining techniques such as misalignment correction, soft labeling, and pseudo labeling in his research of contrail detection using satellite imagery~\cite{Lin2024}. Airborne Object Paper~\cite{Domine_2025}.

\section{Data Collection and Preprocessing}\label{sec:s3}
For this research, we incorporated two Airborne Object Detection datasets. The first one is Aerial Vehicle Detection (AVD) Dataset~\cite{llpukojluct2023aerialvehicle}, which was taken directly from Kaggle. Some useful information about the dataset is presented in Table~\ref{tab:data}. 

The second dataset, named the Airborne Object Detection and Threat Analysis (AODTA)~\cite{3yephw9626}\footnote{\href{https://ieee-dataport.org//documents/airborne-object-detection-and-threat-analysis-aodta-dataset}{AODTA Dataset [Click Here]}} was constructed by integrating four distinct datasets: the Commercial Aircraft Dataset~\cite{nelyg8002000_commercial_aircraft_2025}, the Drones Dataset (UAV)~\cite{hasan2022drone}, the Helicopter dataset~\cite{nelyg8002000_helicopter_dual_rotor_2025}, and the Birds and Drone dataset~\cite{harshwalia_birds_vs_drone_2025}. All the above-mentioned datasets are supporting Creative Commons Public Domain (CC0) licensing, and are permitted to be used for academic research works.

\begin{table}[htbp]
\tiny
\caption{Dataset Information}
\label{tab:data}
\centering
\resizebox{1.0\linewidth}{!}{
\begin{tabular}{lllc} \hline
\textbf{Dataset} &\textbf{Class} &\textbf{Number of Images} &\textbf{After Augmentation}\\ \hline
& Airplane &236 &NA\\
AVD Dataset~\cite{llpukojluct2023aerialvehicle} &Drone &902 &NA\\
& Helicopter & 274  & NA\\
& UAV & 7046  & NA\\ \hline
& \textbf{Total}& \textbf{8458}\\ \hline
&  Airplane~\cite{nelyg8002000_commercial_aircraft_2025} &  6538 & 6538\\
AODTA Dataset~\cite{chatterjee2025airborne} & Drone~\cite{hasan2022drone} & 2194 & 6538\\
& Helicopter~\cite{nelyg8002000_helicopter_dual_rotor_2025} & 1119  & 6538\\
& Birds~\cite{harshwalia_birds_vs_drone_2025} & 428  & 6538\\ \hline
& \textbf{Total}& \textbf{10,279}& \textbf{26,152}\\ \hline
\end{tabular}
}
\end{table}

In AVD dataset~\cite{llpukojluct2023aerialvehicle}, aerial objects were categorized into four classes: Airplane, Drone, Helicopter, and UAV, with threat levels as “Low,” “Medium,” and “High” assigned based on observable features such as type, speed, and weaponry. 

Similarly, AODTA Dataset~\cite{3yephw9626} was designed using four different dataset objects, such as Airplane, Drone, Helicopter, and Birds, with assigned threat levels as "High", "Medium", and "Low" according to the same observable features mentioned above.  

\begin{figure*}[htbp]
\begin{center}
\includegraphics[width=1\linewidth]{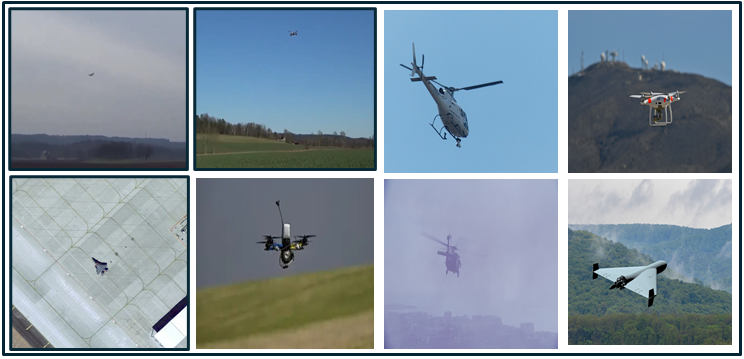}
\caption{No Threat Aerial Object Image from the dataset (a) Airplane Civilian Aircraft, (b) Drone Civilian Model, (c) Helicopter Civilian Aircraft, (d) UAV Civilian Model vs. Threat Aerial Object Images from the Airplane Military Aircraft, (f) Drone Military Model, (g) Helicopter Military Aircraft, (h) UAV Military Model}
\label{fig:sd}
\end{center}
\end{figure*}

Data preprocessing involved cleaning and normalizing the images to maintain consistency in quality and resolution. Data augmentation techniques, including rotation, scaling, and flipping, were applied to improve the dataset's variability and robustness. The images are resized to a target size of (32, 32) with a batch size of 8. We split the data into an 80:20 ratio as train vs test on the AVD dataset and a 70:30 ratio for the AODTA Dataset. Shuffling is applied to the training subset to improve training performance. Figure~\ref{fig:sd} represents some sample images from the dataset~\cite{llpukojluct2023aerialvehicle}. Table~\ref{tab:threat-annotation} represents the threat annotation criteria in the dataset.
\begin{table}[htbp]
\centering
\caption{Examples of Threat Annotation Criteria in the Dataset}
\label{tab:threat-annotation}
\begin{tabular}{|l|l|c|}
\hline
\textbf{Object Type} & \textbf{Example} & \textbf{Threat Level} \\ \hline
Airplane   & Civilian Jet   & Low   \\ \hline
Airplane   & Fighter Jet    & High \\ \hline
Bird & Live Bird    & Low \\ \hline
Bird (Ornithopter) & Military UAV   & High \\ \hline
Drone      & Hobby Drone    & Low  \\ \hline
Drone      & Military UAV   & High  \\ \hline
Helicopter & News Chopper   & Low   \\ \hline
Helicopter & Attack Heli    & High \\ \hline
\end{tabular}
\end{table}

 \begin{figure*}[htbp]
	\begin{center}
		\includegraphics[width=1\linewidth]{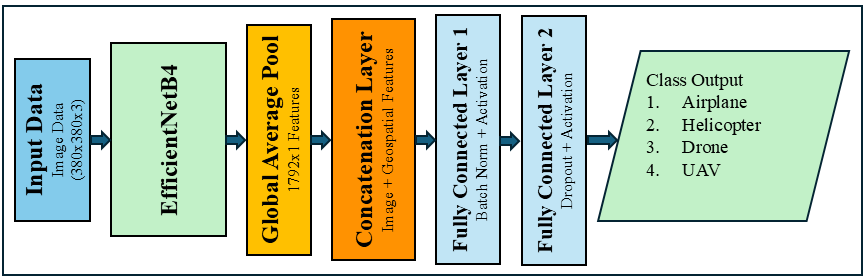}
		\caption{Airborne Object Class Detection Model}
		\label{fig:c1}
	\end{center}
\end{figure*}

\begin{figure*}[htbp]
	\begin{center}
		\includegraphics[width=1\linewidth]{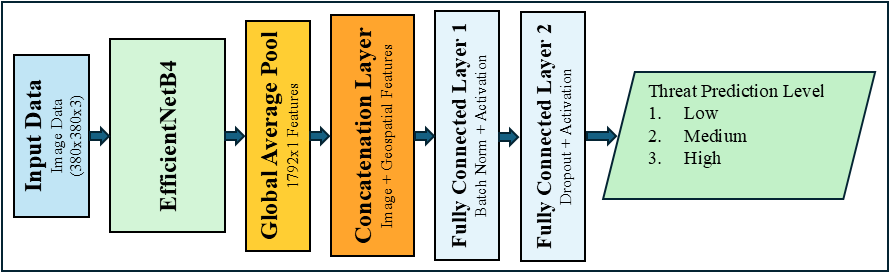}
		\caption{Airborne Object Threat Detection Model}
		\label{fig:c2}
	\end{center}
\end{figure*}
Notably, in this work, we cautiously avoid object detection by utilizing object localization via bounding boxes. Instead, we focus on classification and threat annotation tasks on dataset objects.
\section{Model Architecture}\label{sec:s4}
EfficientNetB4~\cite{TensorFlow_EfficientNetB4}, pre-trained on ImageNet, was chosen for its proven efficiency and accuracy in handling large-scale image classification tasks. The architecture was adapted with two output layers: a softmax-activated “class\_output” layer predicting object types and a softmax-activated “threat\_output” layer assessing threat levels. Intermediate layers featured upsampling, convolution, and global average pooling to optimize feature extraction. A categorical cross-entropy loss function for each task guided the model’s multi-output framework. EfficientNetB4, pre-trained on ImageNet with an input shape of (32, 32, 3), served as the backbone for feature extraction. An upscaling layer followed by a 3x3 convolutional block enhances feature resolution, followed by a 1x1 convolution and global average pooling for refined feature representation. It has two outputs: one predicting object classes and the other determining threat levels, both using softmax activation. The architecture is optimized with categorical cross-entropy loss for each output, and accuracy is used as the evaluation metric. The Adam optimizer is applied with a learning rate of 0.0001.

To improve generalization, the data augmentation pipeline includes the following image techniques: random rotations, width and height shifts, shear transformations, zooming, and horizontal flipping to exhibit variation in aerial scenarios. 
\begin{figure*}[htbp]
	\begin{center}
	\includegraphics[width=1\linewidth]{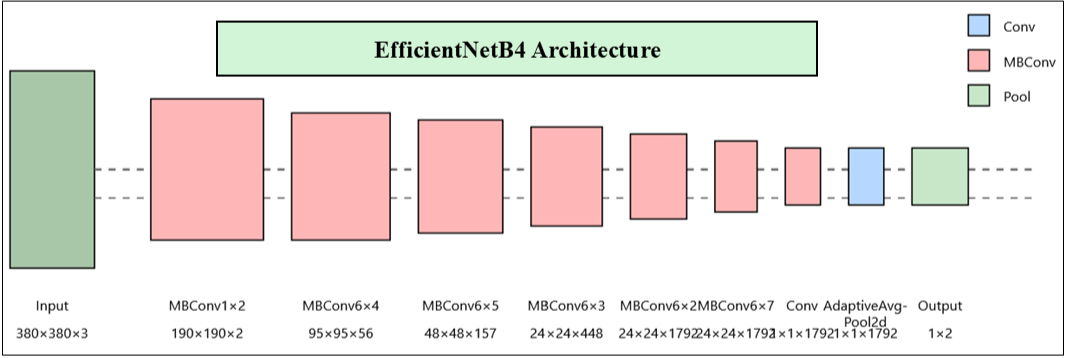}
		\caption{EfficientNetB4 Model Architecture}
		\label{fig:f1}
	\end{center}
\end{figure*}
\section{Experimental Results \& Discussion}\label{sec:s5}
\subsection{Model Performance} 
The EfficientNetB4 model demonstrated strong performance in classification, achieving a class prediction accuracy of 83\% and a threat-level prediction accuracy of 80\% on the AVD Dataset. On the AODTA dataset, it achieved 96\% accuracy on class-level prediction accuracy and 90\% on threat-level prediction accuracy. We compared the performance of EfficientNetB4 with ResNet-50. However, ResNet-50 showed below-average accuracy in both class and threat-level prediction.
\subsubsection{AVD Dataset}
\begin{figure*}[htbp]
	\begin{center}
		\includegraphics[width=0.99\linewidth]{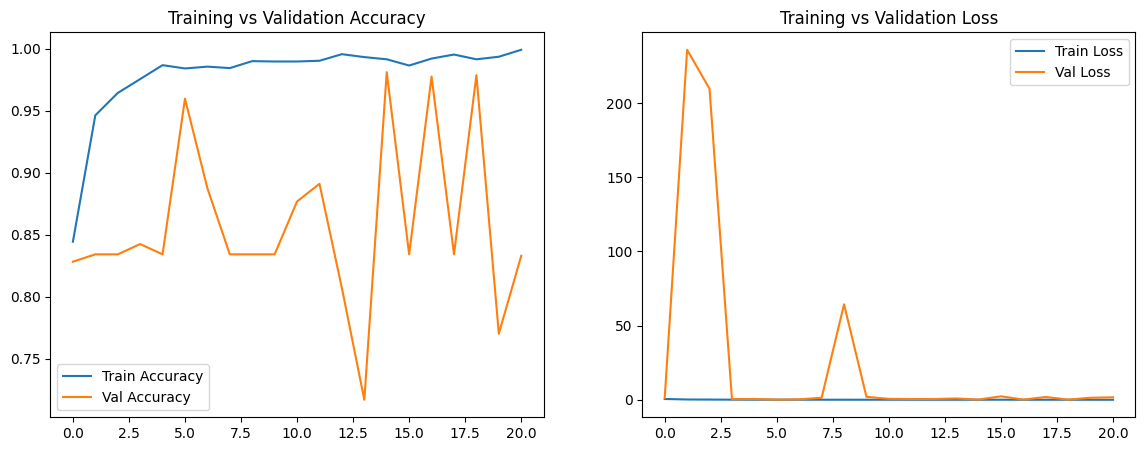}
		\caption{Training vs. Validation Accuracy and Training vs. Validation loss on AVD dataset. Epochs used: 21}
		\label{fig:c3}
	\end{center}
\end{figure*}

Figure~\ref{fig:c3} shows the training accuracy and training loss graphs using the EfficientNetB4 model on the AVD dataset, utilizing 21 epochs.

\begin{figure}[htbp]
	\begin{center}
		\includegraphics[width=1\linewidth]{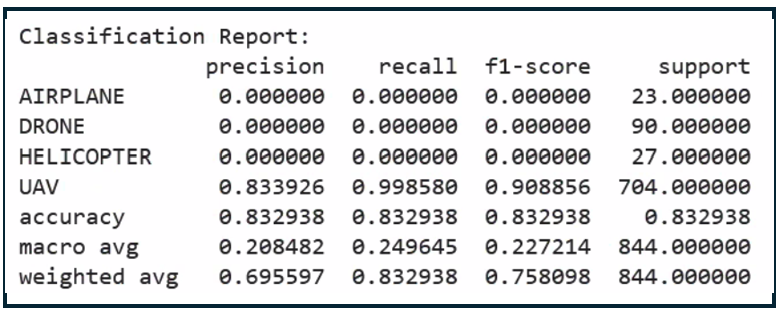}
		\caption{Precision, Recall, F1-Score, and Support for Aerial Object Classification on EfficientNetB4 on the AVD dataset}
		\label{fig:old_f1score}
	\end{center}
\end{figure}

Figure~\ref{fig:old_f1score} demonstrates the classification report, including Precision, Recall, f1-Score, and Support of the EfficientNetB4 on the AVD dataset. 

\begin{figure}[htbp]
	\begin{center}
		\includegraphics[width=0.8\linewidth]{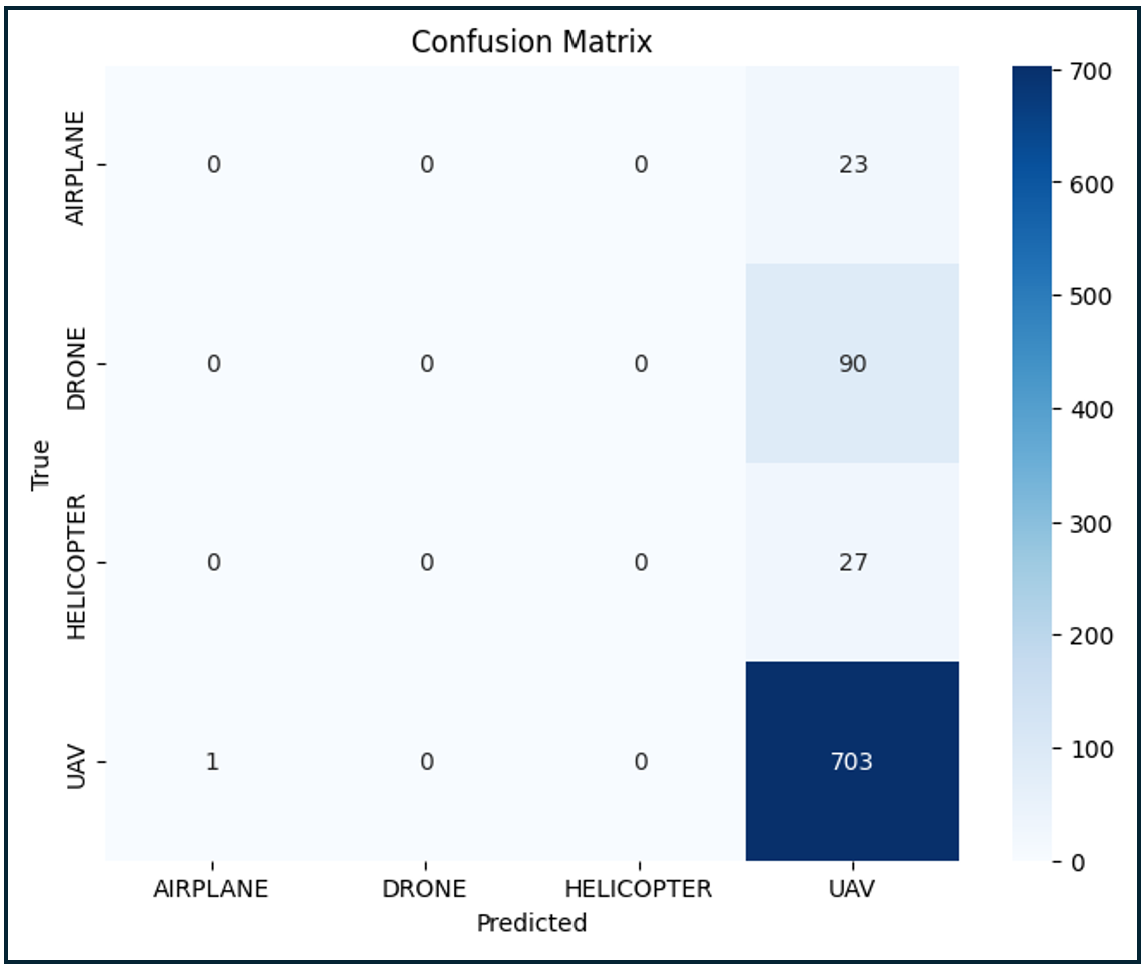}
		\caption{Confusion Matrices on Class Label Prediction on AVD dataset}
		\label{fig:EffB4_conf_old}
	\end{center}
\end{figure}

Figure~\ref{fig:EffB4_conf_old} demonstrates the confusion matrices of the EfficientNetB4 on the 4 classes, using the AVD dataset.  

\subsubsection{AODTA-Dataset}
\begin{figure*}[htbp]
	\begin{center}
		\includegraphics[width=0.99\linewidth]{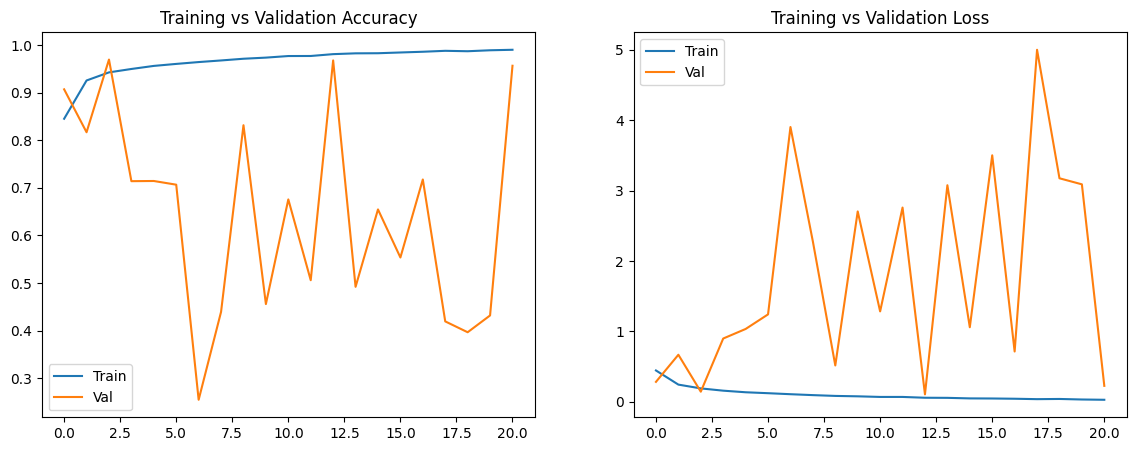}
		\caption{Training vs. Validation Accuracy and Training vs. Validation Loss Graphs for AODTA Dataset. Epochs used: 21}
		\label{fig:hybrid-accu-loss}
	\end{center}
\end{figure*}

\begin{figure}[htbp]
	\begin{center}
		\includegraphics[width=1\linewidth]{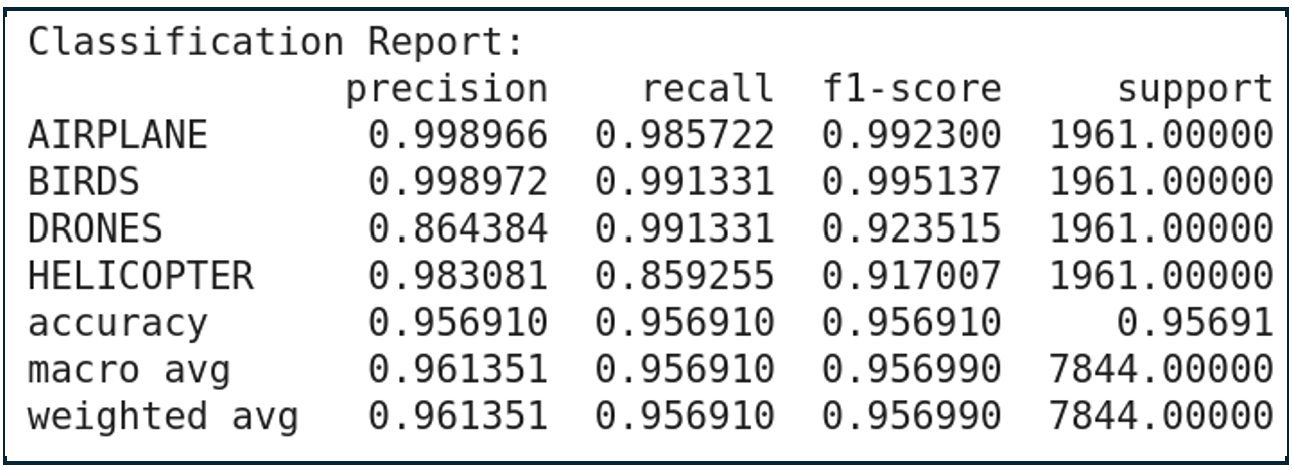}
		\caption{Precision, Recall, F1-Score, and Support for Aerial Object Classification on EfficientNetB4 on the AODTA dataset}
		\label{fig:EffB4_f1score}
	\end{center}
\end{figure}

Figure~\ref{fig:hybrid-accu-loss} shows the training accuracy and training loss graphs using the EfficientNetB4 model on the AODTA dataset using 21 epochs.
Figure~\ref{fig:EffB4_f1score} shows the classification report, including Precision, Recall, f1-Score, and Support of the EfficientNetB4 on the AODTA dataset.

\begin{figure*}[htbp]
	\begin{center}
		\includegraphics[width=1\linewidth]{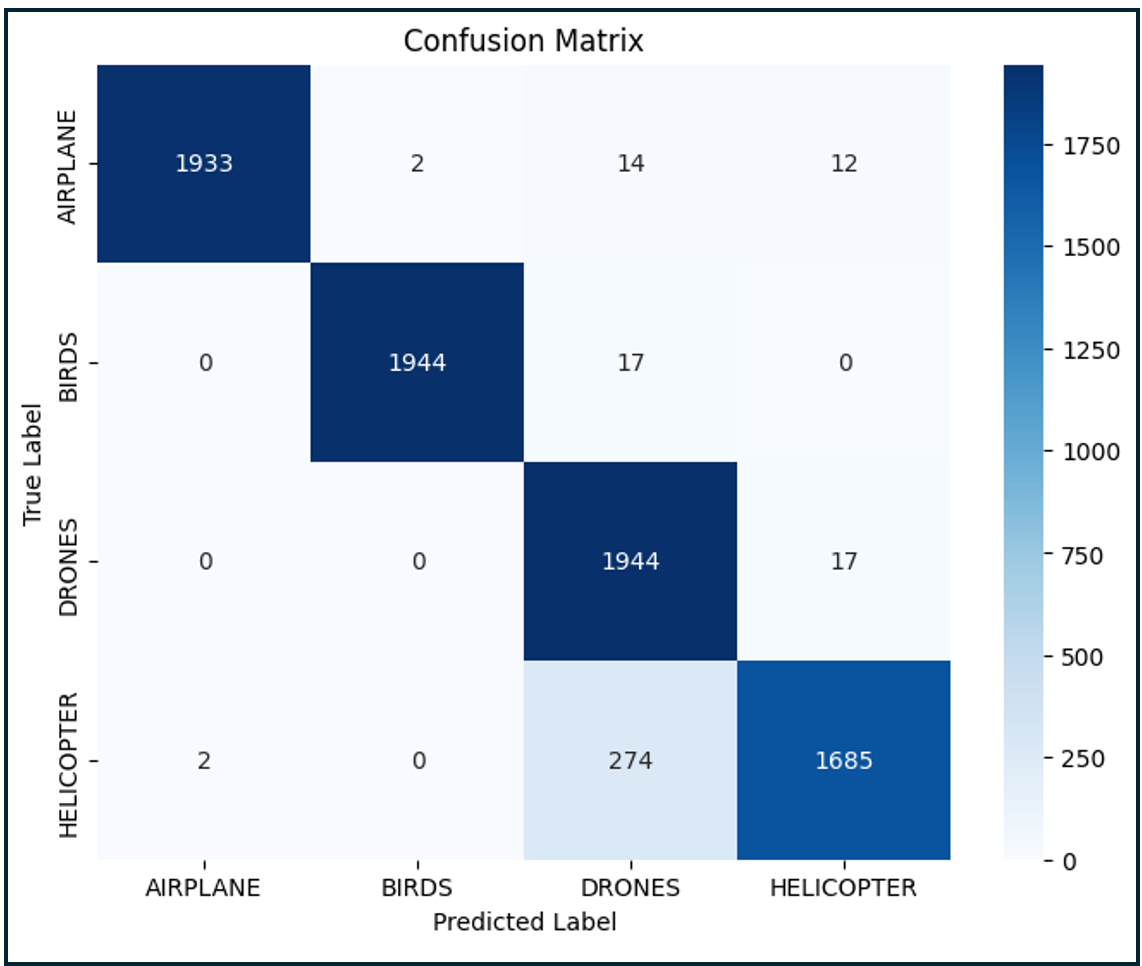}
		\caption{Confusion Matrices on Class Label Prediction on AODTA dataset}
		\label{fig:EffB4_conf_class}
	\end{center}
\end{figure*}

Figure~\ref{fig:EffB4_conf_class} demonstrates the confusion matrices of the EfficientNetB4 on the 4 classes, using the AODTA dataset.  

The AVD dataset, when evaluated using EfficientNetB4~\cite{TensorFlow_EfficientNetB4}, resulted in comparatively lower classification accuracy than our proposed AODTA dataset, as evident from the confusion matrices shown in Figure~\ref{fig:EffB4_conf_class}. Given this, we opted to focus on the AODTA dataset for a more reliable analysis and benchmarked the performance of EfficientNetB4~\cite{TensorFlow_EfficientNetB4} against the widely used ResNet-50 architecture~\cite{Chatterjee2025}. The results indicated that ResNet-50 struggled significantly with accurate classification, exhibiting a high number of false positives and false negatives across all classes—airplanes, birds, drones, and helicopters as seen in the confusion matrix in Figure~\ref{fig:EffB4_conf_class}. In contrast, EfficientNetB4~\cite{TensorFlow_EfficientNetB4} consistently demonstrated better classification precision and robustness, especially in handling complex aerial object features. These findings suggest that airborne object detection tasks—where precision and reduced misclassification are critical—can be performed more effectively and reliably using EfficientNetB4~\cite{TensorFlow_EfficientNetB4} over traditional models like ResNet-50. Therefore, EfficientNetB4~\cite{TensorFlow_EfficientNetB4} offers a more efficient and scalable solution for real-time airborne object detection in surveillance and monitoring applications. 

\begin{figure*}[htbp]
	\begin{center}
		\includegraphics[width=1\linewidth]{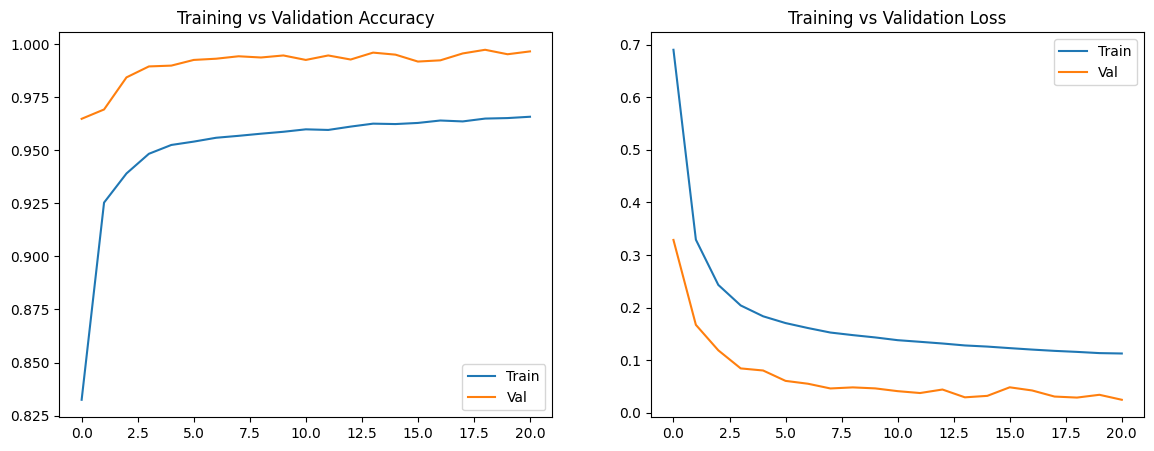}
		\caption{Model Accuracy and Loss Graphs of ResNet-50 on AODTA Dataset. Epochs used: 21}
		\label{fig:resnet-accuracy-loss}
	\end{center}
\end{figure*}

\begin{figure}[htbp]
	\begin{center}
		\includegraphics[width=1\linewidth]{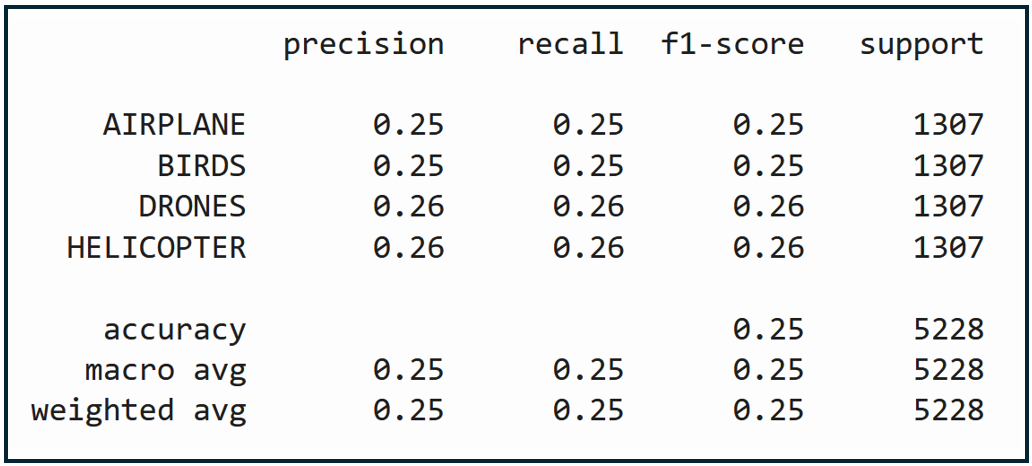}
		\caption{Precision, Recall, F1-Score, and Support for Aerial Object Classification on ResNet-50 on the AODTA dataset}
		\label{fig:resnet_f1score}
	\end{center}
\end{figure}

Figure~\ref{fig:resnet-accuracy-loss} shows the training accuracy and training loss graphs using the ResNet-50~\cite{Chatterjee2025} model on the AODTA dataset using 21 epochs.
Figure~\ref{fig:resnet_f1score} shows the classification report, including Precision, Recall, f1-Score, and Support of the  ResNet-50~\cite{Chatterjee2025} on the AODTA dataset.

\begin{figure}[htbp]
	\begin{center}
		\includegraphics[width=1\linewidth]{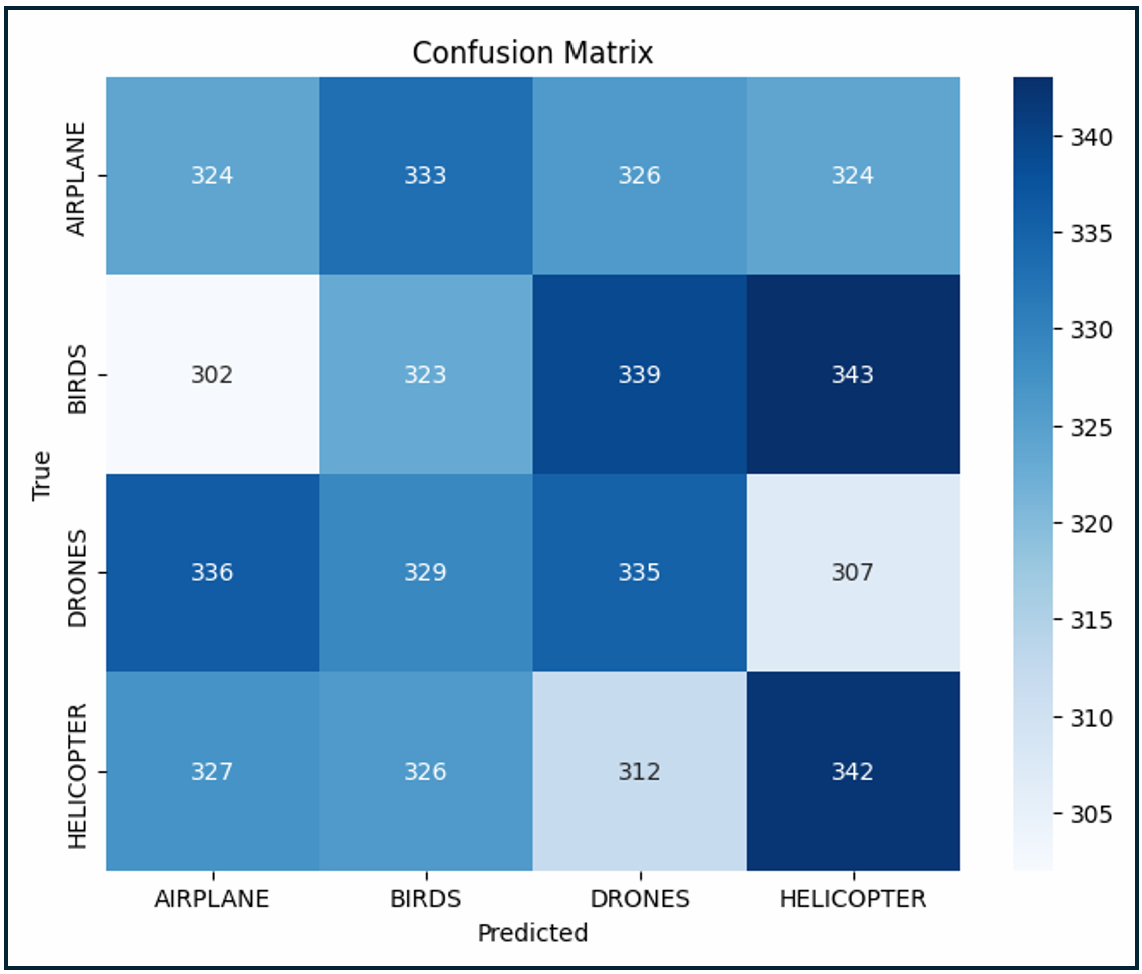}
		\caption{Confusion Matrics on Class Label Prediction of ResNet-50 on AODTA dataset}
		\label{fig:resnet_conf_class}
	\end{center}
\end{figure}

Figure~\ref{fig:resnet_conf_class} demonstrates the confusion matrices of the ResNet-50~\cite{Chatterjee2025} on the 4 classes, using the AODTA dataset.  

\subsection{Discussion and Analysis}
On the class label predictions, precision and recall values for all object classes exceeded 0.9, reflecting balanced performance across categories on the AODTA dataset on EfficientNetB4 as summarized in Figure~\ref{fig:EffB4_f1score}. 

On the threat Level predictions the EfficientNetB4 achieved high precision and recall for high threats, while low and medium threats faced challenges due to dataset imbalance on the AVD dataset as summarized in Figure~\ref{fig:old_f1score}. Notably, the EfficientNetB4 model attained an overall accuracy of 90\%, with the highest precision and recall observed for high-threat categories like UAVs and drones, shown in Figure~\ref{fig:old_f1score} on the AVD dataset and Figure~\ref{fig:EffB4_f1score} on the AODTA dataset.

However, on the AVD dataset, the model showed poor performance for low and medium threat levels due to class imbalance and overlapping features, resulting in zero precision and recall for these categories as shown in Table~\ref{tab:threat_level_classification}.


The training and validation accuracy curves touched by the third epochs as shown in Figure~\ref{fig:hybrid-accu-loss}, indicate effective learning, while the loss curves stabilized, confirming the model’s robustness of EfficientNetB4 as shown in Figure~\ref{fig:c3} on the AVD dataset and Figure~\ref{fig:hybrid-accu-loss} on the AODTA dataset. 

The confusion matrices revealed frequent misclassifications between these categories, as shown in Figure~\ref{fig:EffB4_conf_old}, on the AVD dataset using EfficientNetB4 model, attributed to low image quality and the very small airborne objects captured by a low-resolution camera. 

Overall, our study demonstrates that integrating object classification and threat-level prediction within a single deep learning framework is both feasible and effective. This approach holds strong potential for real-world applications in defense, surveillance, and airspace management.
	\begin{table}[h!] \centering
		\caption{Threat Level Classification Report on AVD Dataset} \label{tab:threat_level_classification}
		 \begin{tabular}{l|c|c|c|c} \hline \textbf{Threat Level} & \textbf{Precision} & \textbf{Recall} & \textbf{F1-Score} & \textbf{Support} \\ \hline 
			LOW & 0.00 & 0.00 & 0.00 & 23 \\  MEDIUM & 0.00 & 0.00 & 0.00 & 27 \\ 
		  HIGH & 0.94 & 1.00 & 0.97 & 794 \\ \hline \textbf{Accuracy} & - & - & - & 844 \\ \hline \textbf{Macro Avg} & 0.31 & 0.33 & 0.32 & 844 \\  
			\textbf{Weighted Avg} & 0.89 & 0.94 & 0.91 & 844 \\ \hline 
			\end{tabular} 
			 \end{table}
\section{Conclusion}\label{sec:s6}
This study introduces a dual‑task framework for aerial object classification and threat‑level prediction built on the EfficientNetB4 architecture. A central contribution of this work is the development of the AODTA Dataset, created to overcome limitations in existing resources by offering cleaner, more diverse, and better‑balanced data for airborne object analysis.

Experimental results demonstrate that EfficientNetB4 consistently outperforms ResNet‑50, particularly on the AODTA Dataset, achieving 96\% accuracy in object class prediction and 90\% accuracy in threat‑level classification. These findings highlight the value of pairing a well‑curated dataset with a high‑capacity model architecture, enabling significant performance gains for surveillance, defense, and airspace management applications.

Future research should explore expanding data diversity through advanced augmentation strategies and synthetic data generation, as well as adapting the framework for real‑time operation in dynamic environments. We also plan to integrate a full detection pipeline to evolve this work into an end‑to‑end detection‑and‑classification system suitable for real‑time deployment.

Overall, this research represents a substantive step toward automated airborne threat assessment and establishes a strong foundation for future advancements in real‑time air defense and security systems.


\end{document}